# Precision Meets Art: Autonomous Multi-UAV System for Large Scale Mural Drawing

Andrei A. Korigodskii[1,2*], Artem E. Vasiunik[1,3], Georgii A. Varin[1,4], Adilia M. Zukhurova[1,4], Matvei V. Urvantsev[1], Semen A. Osipenkov[1], Igor S. Efremov[1,5], Georgii E. Bondar[1,4]

*Abstract* — *The integration of autonomous unmanned aerial vehicles (UAVs) into large-scale artistic projects has emerged as a new application in robotics. This paper presents the design, deployment, and testing of a novel multi-drone system for automated mural painting in outdoor settings. This technology makes use of new software that coordinates multiple drones simultaneously, utilizing state-machine algorithms for task execution. Key advancements are the complex positioning system that combines 2D localization using a single motion tracking camera with onboard LiDAR for precise positioning, and a novel flight control algorithm, which works differently along the trajectory and normally to it, ensuring smoothness and high precision of the drawings at the same time. A 100 square meters mural was created using the developed multi-drone system, validating the system's efficacy. Compared to single-drone approaches, our multi-UAV solution significantly improves scalability and operational speed while maintaining high stability even in harsh weather conditions. The findings highlight the potential of autonomous robotic swarms in creative applications, paving the way for further advancements in large-scale robotic art.*

## I. INTRODUCTION

In recent years, scientific progress have marked the start of a new era in technology, characterized by the integration of creative arts with cutting-edge innovations [1], [2], [3]. This convergence has introduced the use of autonomous unmanned aerial vehicles (UAVs) in the world of art, particularly in graffiti and mural drawing [4], [5], [6]. Despite being a relatively new field of study, multiple projects have already demonstrated success [7], [8], [9].

Previously, our team presented a study on the development of an autonomous painter-drone capable of functioning in outdoor conditions, such as resistance to strong winds and direct sunlight. We showcased its usage in creating the world's largest mural [10]. However, new experience and the latest developments in painting UAVs technology [11], [12], [13] have shown a path for further advancements. Primarily, a multi-drone solution is required to increase the efficiency of large-scale projects [14], [15], [16].

This study focuses on the design of a multi-UAV system for outdoor painting projects. Our team developed a new drone model, introduced a fault-safe battery changing system and added unique IR markers for each drone.

Additionally, the team developed new software that allows multiple drones, controlled by state-machine algorithms, to be controlled from the central control system simultaneously. The combined system was tested during the creation of a 100 m$^2$ mural, where it demonstrated high efficiency and accuracy while drawing collaboratively with a human artist.

The goal of our paper is to highlight the advantages of multi-drone systems in the field of street art and mural drawing, demonstrating their ability to facilitate the quick and efficient creation of large-scale art projects. By exploring the use of these technologies in our project, we aim to encourage further development in this area.

## II. SYSTEM DESIGN

### A. System architecture

The system consists of several unmanned aerial vehicles (UAV), equipped with single-board companion computers, and a ground station with a single camera motion tracking system. Robot Operating System (ROS) was used for communication between the parts of the system. An overview of the system program architecture is presented in Figure 1.

TABLE I. COMPARISON OF LOCALIZATION TECHNIQUES

| Technique | Advantages and disadvantages |
|---|---|
| IMU + Baro | + Does not require additional components<br>– Very high drift makes it unusable |
| GNSS RTK or UWB radio | + Low cost<br>– Quite low precision<br>– Susceptible to multipath errors close to the wall |
| Stationary fiducal markers (onboard recognition) | + Low cost<br>± Intermediate precision<br>– Requires heavy onboard processing<br>– Requires placement of multiple fiducial markers |
| Motion capture | + High precision<br>– Extremely high cost to cover necessary volume |

* Corresponding author, e-mail: akorigod@gmail.com
[1] Sverk Ltd., Moscow
[2] Lomonosov Moscow State University
[3] Cognitive Pilot Ltd., Moscow
[4] NUST MISIS, Moscow
[5] Kosygin State University of Russia, Moscow

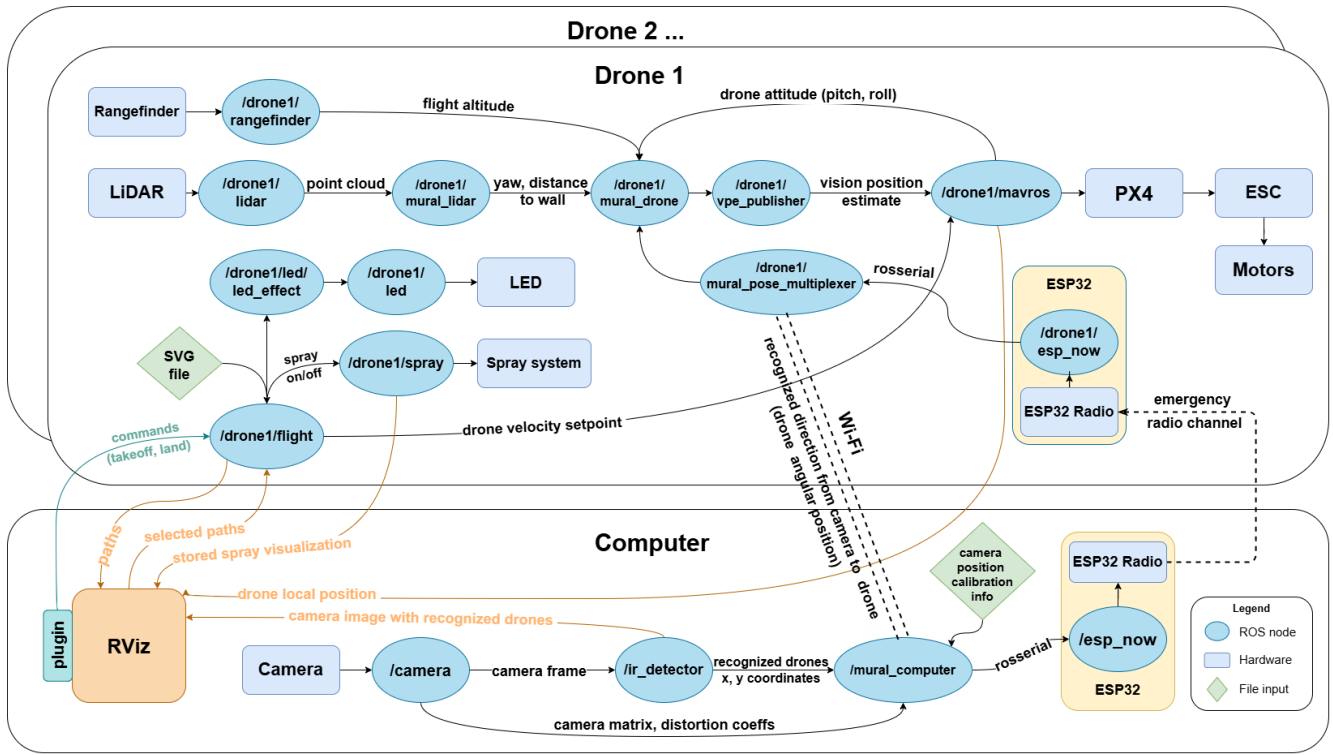

Figure 1. System architecture overview

When deciding on the localization system, our team revised the comparison done last year (refer to Table I). However, none were deemed suitable for project needs, so our team opted to expand on developed last year navigation system, that combined 2D localization with a camera and on-board LiDAR complex.

### B. Ground control station systems

#### 1) 2D localization system

A high-resolution, low-latency camera with an IR-pass filter was used to detect each drone and estimate its height and lateral movement.

Each UAV was equipped with three equally spaced IR markers arranged in a straight-line pattern positioned on a certain degree angle on its back. The algorithm analyzed the camera image, identifying and grouping the three brightest nearby spots. A straight line was then fitted through each group of spots, and the angle of the resulting lines was used to determine the identity of each UAV on the image. Subsequently, for every drone, the algorithm calculated the center point among its visible markers.

To enhance tracking accuracy, each UAV was assigned an approximate area of interest (200 × 200 pixels in the camera image) based on its recorded position from the previous camera frame. If certain IR markers temporarily became invisible for the camera, the algorithm estimated the drone's center point by calculating it using the remaining visible markers within the corresponding area of interest.

Once the center points of all drones were estimated, the algorithm projected a corresponding number of beams in 3D space, originating from the known camera position and passing through the detected center points. This ensured that the exact 3D position of each UAV lay somewhere along its respective beam. Calculated information was transmitted to each drone's on-board computer via ROS-topic.

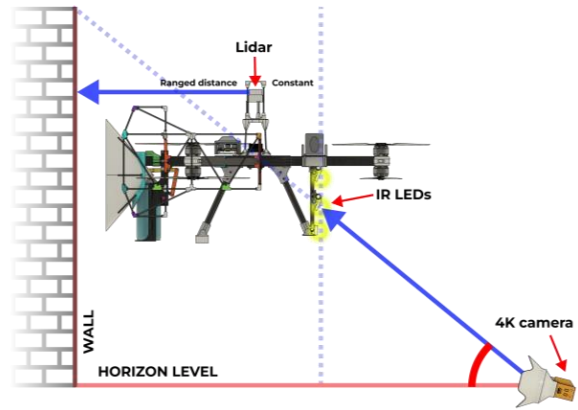

Figure 2. 2D localization system visualization

### C. On-board systems

The following paragraphs describe the computations that were done locally on each drone's respective computer.

#### 1) LiDAR localization

To accurately position itself along the projected beam and determine its 3D coordinates, each UAV required precise distance measurements to the wall. While the onboard flight controller reliably calculated pitch and roll angles, the yaw angle data was subject to rapid drift, making it unreliable. To compensate for this and calculate the missing 3rd coordinate, each drone was equipped with a horizontally aligned 2D LiDAR sensor.

The LiDAR sensor, rotating along the drone's vertical axis, generated a point cloud of the wall. By applying a RANSAC algorithm, the system accurately determined both the yaw angle and the drone's distance from the wall. This enabled precise localization in 3D space by aligning the calculated beam with the corresponding distance measurement.

The final output of the localization algorithm was the precise 3D position of each UAV. This data was transmitted from the onboard computers to each drone's flight control unit via MAVROS and subsequently utilized by the PX4 autopilot to ensure accurate trajectory control.

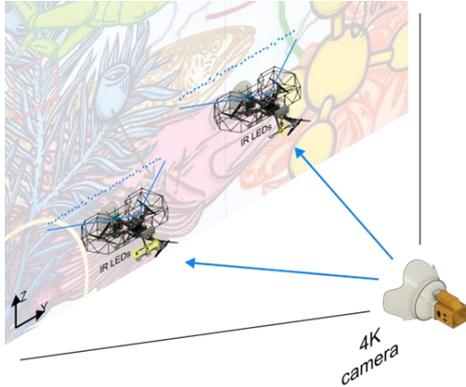

Figure 3. The positioning system combines information from ground control camera and on-board LiDAR

*2) Preprocessing of Flight Trajectories*

To ensure kinematically feasible flight trajectories, some preprocessing steps were required for the SVG images loaded into each drone's computer. Initially, all paths were extended forward and backward by approximately 30 cm. These extensions allowed the drone to reach a stable target velocity before initiating the actual drawing path and decelerating smoothly at the end.

After the SVG image was processed into paths, the following post-processing steps were applied:

- Each line and curve were analyzed, and segments were either joined or separated based on the tangent angle between them.
- Paths shorter than 3–5 cm that were not joined with other paths were removed.

To fill enclosed areas, additional processing was required to generate infill trajectories:

- A bounding box was computed around the target contour.
- A set of horizontal lines was generated within the bounding box at fixed intervals, alternating direction between successive lines.
- Intersections between these lines and the contour were identified to determine valid infill regions.
- To avoid computational errors, intersections below a predefined threshold distance were ignored.
- The resulting lines were sorted, and their direction was adjusted as necessary for complex infills with large internal voids.

All the paths were sorted to minimize non-drawing travel time. Additionally, a bottom-to-top drawing progression was ensured to minimize time in the air.

*D. Trajectory tracking flight control*

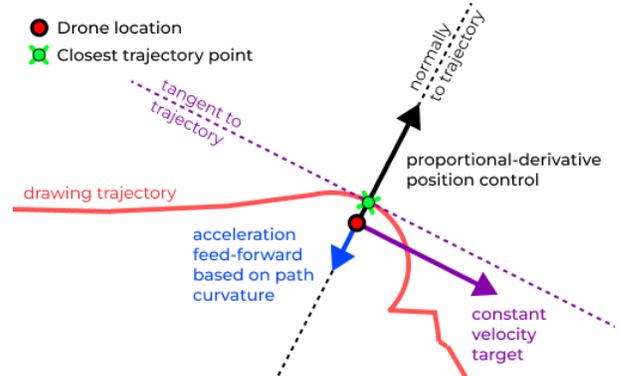

Figure 4. Trajectory tracking control

In order to properly control UAV's flight on a drawing path, high level of precision in flight control was required. To achieve that, control methodologies were different for tangential and normal trajectory tracking. For line drawings maintaining the uniform thickness of the said line was much more crucial than precise position control. Therefore, constant target speed was employed during drawing to eliminate speed oscillations.

On the contrary, precise execution of the position setpoint normal to the trajectory was necessary for maintaining drawing accuracy. This was achieved by using a proportional–derivative (PD) regulator, which output was directly fed into velocity controller. Additionally, another PD regulator was used to maintain the optimal distance to the wall.

*E. Connectivity*

To ensure proper connection between aground station and UAVs, all parts of the system were connected via Ethernet cables to the Wi-Fi router, located near the wall where the mural was created. That helped to maximize the signal strength.

Additionally, a backup link was established, in case of Wi-Fi failure or delays. It consisted of a pair of ESP radio nodules, connected via ESP-NOW protocol. The ground module was connected to the ground station and transmitted the visual navigation data, which was received by the onboard module, connected to the companion computer.

## III. SYSTEM INTEGRATION

This section presents how the developed control system and autonomous flight algorithms were physically integrated in each drone system as shown in Figure 5.

### A. System requirements

To properly execute the project, each drone had to match the following mechanical requirements (refer to Table II).

TABLE II. UAV MECHANICAL REQUIREMENTS

| Lifting capacity | Weight of navigation equipment + 500 g paint canister |
|---|---|
| Wall to spray nozzle distance | 10 cm |
| Structural integrity | Ability to withstand wall or ground collision |
| Propeller air intake | Unobstructed air intake even in the immediate vicinity of the wall |
| Flight stability | Sufficient thrust to withstand harsh weather conditions |

### B. Mechanical design

#### 1) Frame design

The coaxial hexacopter configuration was selected as the optimal drone platform based on the specified requirements. This arrangement minimizes the area interacting with the air, thereby enhancing the stability of the drone. Additionally, the design of this configuration provides a 120-degree angular separation between the arms, allowing for the placement of a can of spray paint closer to the drone's center.

The previous version of the drone frame was based on a heavily modified Tarot 690S hexacopter kit, which was not originally designed for coaxial engines and had significant design flaws, such as 5- to 10-degree rotation of the two-motor group along the axis of one of the frame's beams. To address the issues, the new frame version was constructed using square hollow section pipes, which are fixed between two carbon fiber base plates from the Tarot 690S kit. The new design drastically improved the stability of the motors and reduced the number of components and the overall mass of the drone.

#### 2) Identification system

To enable the system to distinguish between different UAVs, three IR lights were placed on the back of each drone in a unique pattern. These lights were mounted on a carbon tube attached to the drone's frame. The design of the IR marker mount allows for positioning at various angles, facilitating the simultaneous identification of many drones. As an example, for the duration of the project, the first drone had its lights arranged in a horizontal line, while the second drone had them arranged vertically.

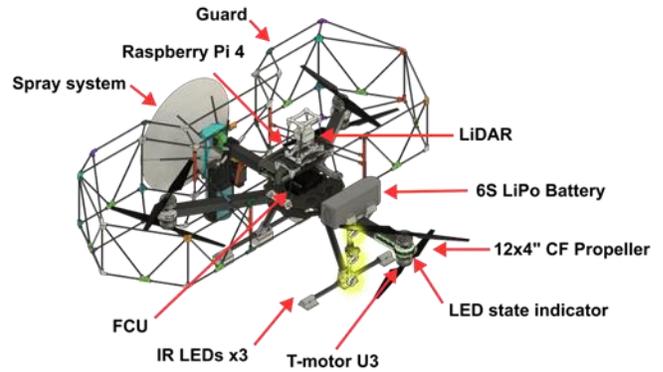

Figure 5. The drone components

#### 3) Battery-Swapping System

A significant design flaw encountered in the previous iteration of the system was the necessity to shut down the drone's systems completely when replacing its battery. This process introduced long delays in the painting operation and could lead to the loss of system files.

Two potential solutions to this problem were suggested: to use pair of supercapacitors that would charge up to 5V or to add a second battery connector and implement forward voltage Schottky diodes to prevent reverse current flow during the batteries swap.

TABLE III. BATTERY SWAPPING SOLUTIONS

| Solution | Advantages and disadvantages |
|---|---|
| Supercapacitors | + Low cost<br>+ Light weight<br>– Limited to 10 seconds swapping time |
| 2 battery connectors with Schottky diodes | + No time limitations<br>– Higher cost<br>– Increased weight |

After careful comparison (refer to Table 3), the team opted to use supercapacitors, which implementation greatly cut down on maintenance time.

#### 4) Paint spraying system

Spray paint dispensation consisted of high force spraying system pressing on the can cap with a delay of 100–200 milliseconds and a carbon frame with a cone made of cardboard, for protection of paint flow from weather and air.

The drone was designed with an intend for it to help artists more efficiently create their works. So, two different types of caps for spraying patterns were created. The first one is a wide cap for flat vertical spraying. Such spray pattern would help artists to fill big outlines fill solid color. Secondly, a thin cap for drawing precise lines. This would be useful for adding details in areas where humans cannot reach without external help.

## IV. SYSTEM OPERATIONS

This section explains the operational procedures of the system

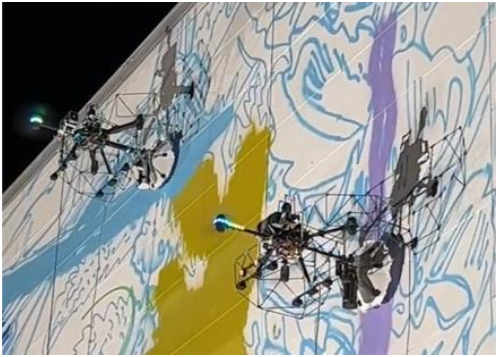

Figure 6. Both drones working simultaneously

### A. System operations basics

The system comprised of a ground control center positioned near the wall where the mural was being created, and two drones. An infrared (IR) camera was installed in proximity to the control center and connected to the main computer, which continuously executed a drone-tracking algorithm. Upon activation, each drone established a connection to the main computer via ROS and initiated a state-machine algorithm that governed its actions.

The system was operated according to the following procedure:

1. SVG file, containing the drawing paths, was uploaded to each drone's onboard computer.
2. Drawing modes and settings were configured via the user interface and stored as a file on the drone, eliminating the need for repeated reconfiguration.
3. The operator selected the SVG paths through the interface, prompting the system to automatically generate and display a preview of them.
4. Upon activation of the takeoff command, the drone autonomously executed the takeoff, positioning, and drawing sequences.
5. The drone continued flight operations and executed the painting task until all selected paths were completed, or until a termination condition was met, such as battery depletion, spray can exhaustion, or manual intervention via the landing command.

In cases of unplanned interruptions during the painting process, each drone automatically saves information about the progress made up to that point. This enables it to later resume painting precisely from where it had stopped. Additionally, adjustments to the drone's settings could be made directly through the graphical user interface without requiring the drone to land.

### B. Graphical User Interface

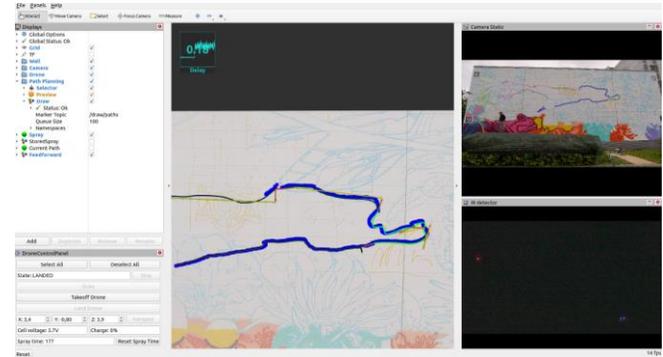

Figure 7. Screenshot of user interface

The graphical user interface (GUI) is made with rqt framework. RViz tool was utilized to show drawing trajectories and UAV's positions and for reprojection of the camera image onto the wall in. This allowed the operator to accurately detect misalignments in drones' trajectories between the real-life flights and planned paths. This was especially important for this project, since sometimes UAVs were drawing simultaneously with a human artist, so all the lines had to be perfectly aligned.

A unique namespace was assigned to each drone, enabling the simultaneous control of multiple UAVs from the same graphical interface. GUI provided real-time telemetry data for each drone, such as its current operational state, battery charge level, and time passed since the beginning of spray painting. Additionally, some control options for each drone were given to the operator: selection of the operational state, pausing the current action, navigation to specified coordinates, FCU reboot, and buttons for manual drawing start, take-off and landing.

### C. Position tracking algorithm and calibration

To determine the camera's position relative to the wall, a daily calibration procedure was conducted. Four large ArUco markers were precisely placed on the mural wall. At the beginning of each day, a calibration image was captured by the camera with the IR-pass filter removed. The camera's position and orientation were then computed, by identifying markers known positions within the image taken.

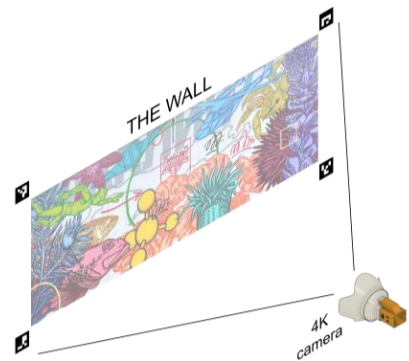

Figure 8. The calibration procedure

## V. Conclusion

The design and deployment of a mural-drawing system for multiple autonomous UAVs is presented in this study, along with an explanation of its working principles during the mural-making process. Enhancing the efficiency of the mural-drawing process was the main goal, and this was accomplished by developing a new user graphical interface, improving the navigation system, refining the UAV frame design, and—most importantly—integrating multiple simultaneously operating drones.

Even though the existing approach for creating big scale murals has proven to be quite successful and simple to execute, there is still room for improvement. To lessen the operator's workload by removing the potential need for manual trajectory corrections to avoid UAV collisions, future research will concentrate on integrating drone-swarm technologies. To minimize the moment of inertia, efforts will also be undertaken to further lower the weight of each drone and position the spray canister closer to the center of mass. Finally, to perhaps break the existing world record for the largest drone-painted mural, our team plans to launch an even more ambitious project that will involve a higher number of UAVs flying concurrently.

We think such large-scale initiatives are not only possible but can be widely implemented globally, making high-quality street art more accessible, especially considering the level of technological maturity attained in our system.

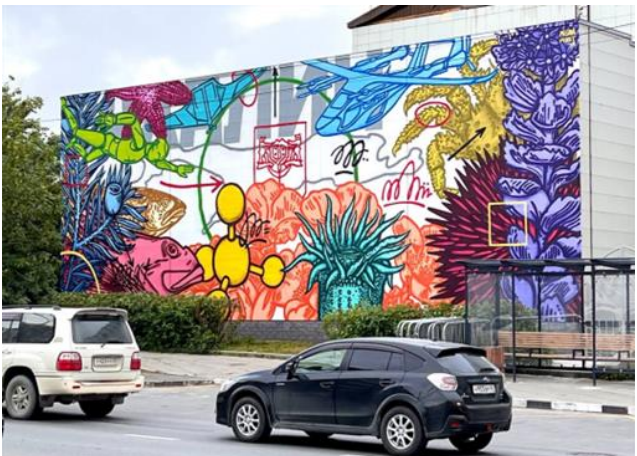

Figure 9. The resulting mural

## Acknowledgement

The authors would like to thank Universal University and 20.35 University for their financial and organizational support, which was instrumental for this study.

The authors thank the artistic director of the project, Misha Most, for his invaluable help and artistic guidance throughout the project, as well as for creating the mural in collaboration with our drones.